\pdfoutput=1

\documentclass[11pt]{article}

\usepackage[final]{acl}

\usepackage{times}
\usepackage{latexsym}

\usepackage[T1]{fontenc}

\usepackage[utf8]{inputenc}

\usepackage{microtype}

\usepackage{inconsolata}

\usepackage{graphicx}

\usepackage{amsmath}
\usepackage{amssymb}
\usepackage{booktabs}
\usepackage{cleveref}
\usepackage{enumitem}
\usepackage{adjustbox}

\newcommand{\dataset}{When2Call}
\newcommand{\genmodel}{Mixtral 8x22B}
\newcommand{\genmodellong}{Mixtral 8x22B \citep{mistral_mixtral8x22b_2024}}

\title{When2Call: When (not) to Call Tools}

\author{Hayley Ross\thanks{Equal contribution.}\thanks{Work done while at NVIDIA.} \\
  Harvard University \\ %
  \texttt{hayleyross@g.harvard.edu} \\\And
  Ameya Sunil Mahabaleshwarkar$^{*}$ \\
  NVIDIA \\
  \texttt{ameyasunilm@nvidia.com} \\\AND
  Yoshi Suhara \\
  NVIDIA \\
  \texttt{ysuhara@nvidia.com} \\}

\begin{document}
\maketitle

\begin{abstract}
Leveraging external tools is a key feature for modern Language Models (LMs) to expand their capabilities and integrate them into existing systems.
However, existing benchmarks primarily focus on the accuracy of tool calling---whether the correct tool is called with the correct parameters---and less on evaluating {\em when LMs should (not) call tools}. 
We develop a new benchmark, \dataset{}, which evaluates tool-calling decision-making: when to generate a tool call, when to ask follow-up questions and when to admit the question can't be answered with the tools provided.
We find that state-of-the-art tool-calling LMs show significant room for improvement on \dataset{}, indicating the importance of this benchmark. 
We also develop a training set for \dataset{} and leverage the multiple-choice nature of the benchmark to develop a preference optimization training regime, which shows considerably more improvement than traditional fine-tuning.
We release the benchmark and training data as well as evaluation scripts.\footnote{\url{https://github.com/NVIDIA/When2Call}
}
\end{abstract}

\section{Introduction}

Tool-calling is an increasingly important capability for modern LMs as it allows them to connect with existing APIs or tools to use real-time information, %
retrieve information from databases,
or carry out actions by integrating with existing systems.
This is especially important given recent advances in small language models which can be deployed on devices, as smaller models do not store as much knowledge as larger models, and thus benefit greatly from access to external tools.
In a typical setup, the model is provided with a list of tool or API specifications in the system prompt. %
The model can access these tools by generating one or more tool calls as code (usually JSON) which conforms to the API specification. Any tool calls are intercepted and executed, and the result is returned to the model. In the second step, the model generates a text response to the user based on the tool call result, similar to RAG. 

\begin{figure}[t]
\includegraphics[width=0.49\textwidth]{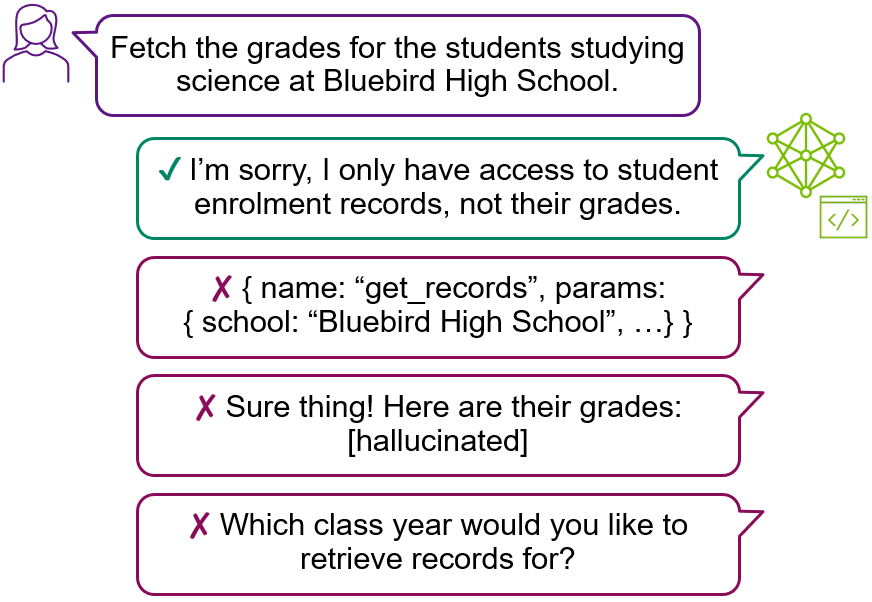}
    \caption{%
    Example of the type of question in \dataset{}. 
    Tool-calling LMs should avoid hallucinating tools or information when given questions they cannot answer.}
    \label{fig:when2call_example}
\end{figure}

\begin{table*}[t]
    \centering
    \begin{tabular}{lccccc} \toprule
        Feature & \dataset{} & BFCL & ToolSandbox & ToolBeHonest & Older \\
        \midrule
        Tool(s) provided, one correct  & $\checkmark$ & $\checkmark$ & $\checkmark$ &  $\checkmark$ &  $\checkmark$  \\
        Tool(s) provided but none correct & $\checkmark$ & $\checkmark$ & $\checkmark$ & $\checkmark$\\
        No tools provided & $\checkmark$ &  &  &  \\
        Question missing information & $\checkmark$ & $\checkmark$ & $\checkmark$ &  \\
        Tool call validation & & $\checkmark$* & \textasciitilde$\checkmark$* & &  $\checkmark$* \\
        Quantifies answer hallucinations & $\checkmark$ &  &  &  \\
        Quantifies tool hallucinations & $\checkmark$ &  & & \textasciitilde$\checkmark^\dagger$ \\
        Quantifies parameter hallucinations & $\checkmark$ &  &  &  \\
        Quantifies follow-up questions & $\checkmark$ &  & &   \\
    \bottomrule
    \end{tabular}
    \caption{Key characteristics of \dataset{} compared to BFCL \citep{yan_berkeley-v1_2024}, ToolSandbox \citep{lu_toolsandbox_2024} and ToolBeHonest \citep{zhang_toolbehonest_2024}, three recent benchmarks. ``Older'' summarizes older benchmarks (see \Cref{sec:related-work}).  
    * Tool calls are validated implicitly by system state in ToolSandbox, in the multi-turn category of BFCL, and some older benchmarks.
    $^\dagger$ ToolBeHonest measures tool hallucination implicitly by solvability judgments.
    }
    \label{tab:comparative_table}
\end{table*}

Notably, the tools provided in the system prompt may differ between train and inference time. %
If a model is not deployed with a tool that answers the question (even if it has seen such tools during training), such as a customer service LM being asked about tomorrow's weather, the LM should say that it cannot answer the question, not hallucinate a previously seen weather tool (or hallucinate tomorrow's weather). 
More subtly, an LM with access only to a database of student records might be asked to retrieve their grades instead, as shown in \Cref{fig:when2call_example}.
A further opportunity for hallucination 
is that the appropriate tool may be available, but the user may not provide enough information to fill its particular required parameters. In this case, we expect the LM to ask a follow-up question, not hallucinate the missing parameters.

The primary focus of most current benchmarks, however, including the current standard, BFCL \citep{yan_berkeley-v1_2024}, is the case where the correct tool is provided and the user provides enough information to call this tool. The benchmark then evaluates whether the correct tool was called with the correct parameters. %
While BFCL's Irrelevance category and ToolSandbox \citep{lu_toolsandbox_2024} consider the cases when the correct tool is not provided or not enough information is provided, they only check if the model generates a tool call or not. Neither evaluates what the model does instead.

We create a new benchmark, \dataset{}, which fills these gaps by explicitly asking the model to choose in a multiple-choice format between four types of behavior: generating a tool call, asking for more information, saying it's unable to answer, or answering the question directly (which amounts to hallucinating the answer, since our questions cannot be answered without tools). 
We summarize the key features of \dataset{} in \Cref{tab:comparative_table}.
We provide both a classical offline multiple-choice evaluation using log-probabilities as well as an LLM-as-judge alternative for closed-source models.

We find that modern tool-calling LMs of all sizes have much room for improvement on \dataset{}, which is unsurprising given that most publicly available training datasets contain many examples of calling tools when tools are provided in the system prompt but a considerably lesser amount of examples of not calling tools when tools are provided.
To address this, we develop a matching training dataset for \dataset{}. We leverage its multiple-choice format to implement supervised fine-tuning (SFT) as well as reward-aware preference optimization (RPO) training \citep{nvidia2024nemotron4340btechnicalreport} and show that RPO training in particular substantially increases performance on \dataset{} benchmark and on BFCL Irrelevance while still maintaining competitive scores on the portion of BFCL where a tool should be called.

\section{When2Call}

\subsection{Tool-calling as Multiple-Choice}

We formulate \dataset{} using a multiple-choice format among behavior types, similar to commonly used LM benchmarks such as MMLU \citep{hendrycks_measuring_2020}.
Specifically, \dataset{} consists of questions with the following four types of answers, as illustrated in \Cref{fig:when2call_example}:
\begin{enumerate}[label=(\alph*)]
  \setlength{\parskip}{0cm}
  \setlength{\itemsep}{0cm}
\item Direct text answer (no tool call)
\item Tool call
\item Follow-up question
\item Unable to answer
\end{enumerate}

We ensure that all questions in \dataset{} require tool use to answer (by requiring real-time information, referring to a database, or similar), such that the direct answer (a) is always a hallucination. 
This allows us to evaluate whether an LM prefers to hallucinate an answer rather than admit that it can't answer the question if no appropriate tool is provided, similar to refusal evaluation \citep{wen_art_2024}.

Using multiple-choice lets us focus on the type of behavior of the model---direct (text) answer, tool call, follow-up question, or ``unable to answer''---rather than having to parse the tool call or classify a generated text answer. We explore classifying generated answers as an alternative in \Cref{sec:llm-as-judge}. 
Multiple-choice approaches have the major benefit of being reproducible and being fast to evaluate. While LLM-as-judge is becoming increasingly common for benchmarks (e.g., MT-Bench; \citealp{mtbench}) so that they can evaluate freely generated responses, scores can change depending on the judge used, and costs can scale quickly.
Further, parsing and evaluating the tool call is already covered by benchmarks like BFCL.
We intend \dataset{} to be complementary to BFCL.

\begin{table*}[t]
\small
    \centering
    \begin{tabular}{l |cccc|ccc|ccc} \toprule
        & \multicolumn{4}{c}{\textbf{Correct answer}} & \multicolumn{3}{|c|}{\textbf{Tools provided}} & \multicolumn{3}{c}{\textbf{Tool requirement}} \\
        \textbf{Dataset split} & (a) & (b) & (c) & (d) & 0 & 1 & 2+ & real-time & database & other \\
        \midrule
        \dataset{} Test & 0 & 1,295 & 1,062 & 1,295 & 258 & 712 & 2,682 & 2,178 & 534 & 940 \\
        ~~~~~~LLM-as-judge subset & 0 & 100 & 100 & 100 & 18 & 61 & 221 & 169 & 46 & 85 \\ \midrule
        \dataset{} Train & 0 & 2,000 & 2,000 & 2,000 & 617 & 2,934 & 2,449 & 2,856 & 1,211 & 1,933 \\
        ~~~~~~Preference training variant & 0 & 4,500 & 3,000 & 3,000 & 918 & 6,860 & 2,722 & 5,092 & 2,096 & 3,312 \\
    \bottomrule
    \end{tabular}
    \caption{Statistics of \dataset{} by correct answer type and the number of tool specifications provided. Answer key: (a) direct answer, (b) tool call, (c) follow-up question, and (d) unable to answer.}
    \label{tab:statistics}
\end{table*}

\subsection{Data Generation} \label{sec:data-pipeline}

We synthetically generate the multiple-choice options and new questions for \dataset{} by building off the \texttt{Simple} and \texttt{Multiple Function} categories of BFCL v2 Live \citep{mao_bfcl-v2_2024} for the \dataset{} benchmark and the \texttt{Simple} and \texttt{Multiple Function} categories of APIGen for the \dataset{} training set. 
We choose the Live (v2) subset of BFCL because it is generated by humans, rather than synthetic, and is permissively licensed. 
This allows us to inherit the correct tool calls for each question, as well as the diversity of each of these datasets across a wide variety of subject domains (see \Cref{sec:dataset-statistics}).

We synthetically generate the new data for \dataset{} in two main steps, using \genmodellong{} for all classification and data generation. Step 1 filters BFCL Live or APIGen \texttt{Simple} and \texttt{Multiple Function} by classifying whether the questions require tool-calling to be answered (such as requiring real-time information or access to a database), or whether they could in principle be answered by a sufficiently knowledgeable pretrained model. We prompt \genmodel{} for this classification; the full prompt is shown in \Cref{tab:classification_prompt} in \Cref{sec:synth-prompts}. 

For each filtered BFCL or APIGen question, Step 2 then generates three questions in \dataset{}: one where the question is unchanged and the tool call from BFCL/APIGen is the correct answer, and two others where the question is modified such that either a follow-up question requesting more information or ``unable to answer'' is the correct answer. For each, we also generate the other three (incorrect) multiple-choice answers. Notably, generating these other answers for the training dataset as well as the benchmark allows us to implement preference optimization training (see \Cref{sec:dpo-training}). The exact prompts are provided in \Cref{sec:synth-prompts}.

Prompts were developed iteratively, with manual quality checking after each iteration. We found that breaking the problem down into steps wherever possible (as described below for follow-up questions) yielded the highest quality results, alongside providing a detailed list of mistakes to avoid. We discuss the quality issues associated with synthetic data generation and the steps we took to avoid them in more detail in \Cref{sec:quality}. 

To generate questions that the model should not be able to answer, we provide \genmodel{} with the tool specification, including the tool's text description, and ask it for a related question that cannot be answered by this tool. %
We ask for a \emph{related} question in order to generate close and thus more difficult mismatches between the question and provided tool(s) than typical for the BFCL Irrelevance category -- see \Cref{sec:tool-mismatch-difficulty}.

To generate questions that require a follow-up question, we parse the tool specification corresponding to the correct tool call and select one required parameter to drop. 
We then instruct \genmodel{} to rewrite the user question to omit the information corresponding to that parameter and to write a follow-up question asking for that parameter. 
By breaking down the problem, we can avoid the generation model needing to parse the tool specification at all and greatly improve the quality and consistency of the generated data.
We apply this only to tool specifications that have at least two required parameters in order to avoid ``empty'' questions (such as ``What is the current stock price?'', where the only required parameter, the stock ticker, has been omitted), resulting in slightly fewer questions of this type.%

\subsection{Statistics} \label{sec:dataset-statistics}

\Cref{tab:statistics} shows the proportions of answer types in each split of \dataset{}.
As discussed in more detail in \Cref{sec:dpo-training}, we include a higher proportion of questions where tool-calling is correct in the RPO training dataset to avoid over-conservativeness. 
\Cref{tab:statistics} also shows the proportions of single vs.~multiple tool specifications. As in BFCL, we expect the questions with multiple tools to be more difficult since the LM has to determine whether any of the tools answer the question. 
Our ratios mirror the ratios between the BFCL Live and APIGen \texttt{Simple} and \texttt{Multiple Function} categories, with the addition of the zero tool category. For questions where a follow-up question is the correct answer, we filter out samples from BFCL Live where the correct tool does not have any required parameters, resulting in fewer samples for this category. This ensures that the synthetically modified questions for the ``follow-up'' category always have a required parameter that is missing.

Finally, \Cref{tab:statistics} reports the type of question, arising from the tool requirement classification in \Cref{sec:data-pipeline}: in what way does this question require a tool? Real-time information forms the largest single category; other categories include database access and specialized tools (such as population modeling), as per the classification in \Cref{tab:classification_prompt}.
We inherit the diverse set of domains from BFCL Live and APIGen  (database access, food ordering, real-time weather, etc.).

\subsection{Difficulty of Tool Mismatches} \label{sec:tool-mismatch-difficulty}

In addition to analyzing how LMs should respond to question/tool mismatches in more detail, \dataset{} also distinguishes itself from BFCL Irrelevance by how difficult many of the question/tool mismatches are, by design.
The provided tools in BFCL Irrelevance are often largely unrelated to the question, making it easy for the LM to tell that they do not match \citep{lu_toolsandbox_2024}. 
This reflects some real-world scenarios, such as asking a customer service LM about the weather, but not others: to distinguish the student records from grades in \Cref{fig:when2call_example}, the LM needs to make a much more subtle judgment.
We address this in \dataset{} firstly by constructing the questions with the target ``unable to answer'' to be in the same semantic domain as the tool. %
Secondly, questions targeting the follow-up question answer are necessarily a close match to the tool, differing only by the absence of a required parameter.

\section{Multiple-Choice Evaluation}

We implement the multiple-choice evaluation using log-probability over the four possible answers to determine the model's choice, rather than having the model generate the choice number (e.g., ``Answer: (a)''). For tool-calling, the meta-task of selecting among answers may be unnatural for the models, and presentation details such as answer order \citep{pezeshkpour-hruschka-2024-large,llm-mcq-bias,gupta2024changinganswerorderdecrease} and the number of answers \citep{rodriguez2005three} can artificially affect accuracy. Log-probabilities allow us to bypass all of this. We report accuracy, (byte-)length-normalized accuracy\footnote{\scriptsize \url{https://blog.eleuther.ai/multiple-choice-normalization/}} and F1 metrics. We implement our multiple-choice evaluation as a task in LM Evaluation Harness \citep{eval-harness}, including the preprocessing necessary for evaluating all models in Table \ref{tab:training-results} (see \Cref{sec:prompt-templates}), which can easily be extended to other models. %

\subsection{Model-Specific Prompt Templates} \label{sec:prompt-templates}

Since every tool-calling model has their own preferred tool-calling syntax \citep{carrigan_tool_2024}, we provide the tool calls as JSON by default but implement custom preprocessing for each model, which provides the system prompt that the model expects for tool-calling and formats the tool specifications as well as the tool call option (b) the way the model expects. This avoids artifacts in the answer log-probabilities from unexpected tool syntax. An example is given in \Cref{sec:system-prompts}.
A similar approach is implemented for BFCL, which has custom model handlers that parse each model's tool call output into the format the evaluation code expects, allowing each model to respond in its preferred way.
For models that do not specify a preferred tool-calling prompt, such as Llama 3.1 8B Instruct, we provide a minimal system prompt describing tool use, shown in \Cref{sec:system-prompts}, and provide the tool call answer in JSON.

\subsection{Results for Community Models}

\begin{table*}[t]
    \centering
    \begin{tabular}{l ccc cc} \toprule
        \textbf{Model} & \multicolumn{3}{c}{\textbf{\dataset{}}} & \textbf{BFCL AST} & \textbf{BFCL Irr.} \\
         & F1 $\uparrow$ & Acc-Norm $\uparrow$  & Tool Hall\% $\downarrow$ & Acc $\uparrow$  & Acc $\uparrow$   \\
        \midrule
        Llama 3.2 3B Instruct & 17.9 & 46.5\% & 52\% & 37.6\% & 46.6\% \\
        Llama 3.1 8B Instruct & 16.6 & 44.2\% & 67\% & 51.6\% & 40.0\% \\
        Llama 3.1 70B Instruct & 37.8 & 46.1\% & 57\% & \underline{68.3}\% & 36.5\% \\
        Qwen 2.5 3B Instruct & 29.8 & 48.9\% & 23\% & 54.8\% & 53.1\% \\
        Qwen 2.5 7B Instruct & 32.0 & 50.9\% & 21\% & 64.1\% & 51.4\% \\
        Qwen 2.5 72B Instruct & 32.8 & 49.2\% & 23\% & \textbf{69.3\%} & 61.1\% \\
        xLAM 7B FC-R & 31.5 & 42.7\% & 24\% & 58.3\% & \textbf{79.8\%} \\
        xLAM 8x22B R & 34.3 & 48.3\% & 9.0\% & 74.7\% & 75.2\% \\
        \midrule
        MNM 4B SFT (baseline) & 29.7 & 47.8\% & 16\% & 57.9\% & 41.1\% \\
        MNM 4B \dataset{}-SFT & 48.1 & 67.8\% & 4.3\% & 51.7\% & 67.5\% \\
        MNM 4B \dataset{}-RPO & \underline{51.0} & \underline{69.1\%} & \underline{1.9\%} & 54.0\% & 77.4\% \\
        MNM 8B SFT (baseline) & 31.9 & 49.1\% & 19\% & 62.2\% & 36.3\% \\
        MNM 8B \dataset{}-SFT & 49.4 & 68.2\% & 7.0\% & 57.5\% & 61.0\% \\
        MNM 8B \dataset{}-RPO & \textbf{52.4} & \textbf{70.0\%} & \textbf{1.2\%} & 62.5\% & \underline{78.1\%} \\
    \bottomrule
    \end{tabular}
    \caption{Results on \dataset{}, BFCL v2 Live AST and BFCL v2 Irrelevance for community tool-calling models, and for our Mistral-NeMo-Minitron models with and without training on \dataset{} using SFT and RPO. 
    For \dataset{}, we show Macro F1, length-normed accuracy, and the tool hallucination rate when no tools are provided (lower is better $\downarrow$; see \Cref{sec:hallucination-calculation} for calculation).
    Models not trained on \dataset{} show much room for improvement; RPO training yields the greatest benefits. The \textbf{best} and \underline{second-best} scores are bolded and underlined.
    }
    \label{tab:training-results}
\end{table*}

We evaluate a range of community models of varying sizes with tool-calling capabilities: Llama 3.1 \citep{dubey2024llama3herdmodels}, Llama 3.2 \citep{llama3.2}, Qwen 2.5 \citep{qwen2.5} and xLAM \citep{zhang2024xlamfamilylargeaction}. Scores are shown in \Cref{tab:training-results}. We report results on the v2 Live portion of BFCL to most closely match our dataset, which is generated from BFCL v2 Live.
We find that performance is far from the ceiling on \dataset{} and does not necessarily improve with model size (e.g., Qwen 2.5 3B/7B/72B). More research is needed to understand this interesting result, which may depend on the training data of each of the model sizes.

In particular, most community models are unwilling to admit they cannot answer the question. This results in low accuracy on the ``unable to answer'' category (see confusion matrices in \Cref{sec:confusion-matrices}), as well as higher tool hallucination rates. We define tool hallucination to occur when the model chooses the tool call answer even though no tool specifications were provided for that question -- in other words, the model hallucinated the specification for the tool it chose in its answer (see \Cref{sec:hallucination-calculation} for more details).
Unwanted tool calls also occur for many of the questions where a follow-up question would be correct, suggesting an over-eagerness to call tools. 
This likely reflects that these models are too specialized for the case when tool-calling is the correct choice and do not see enough (or possibly any) training data involving follow-up questions or admitting inability to answer, as judged by the distributions of current publicly available training datasets \citep{liu_apigen_2024,interstellarninja_hermes_2024,glaiveai_glaive-function-calling-v2_2024}. 

\subsection{Training}
We fine-tune and align Mistral-NeMo-Minitron 4B Base
and 8B Base\footnote{\tiny \url{https://huggingface.co/nvidia/Mistral-NeMo-Minitron-8B-Base}} models \citep{sreenivas2024llmpruningdistillationpractice} using NeMo-Aligner \citep{shen2024nemoalignerscalabletoolkitefficient}. 
Training was carried out on eight NVIDIA 8xH100 GPU nodes, taking approximately 3-4 hours per model.
We show results for three cases: (1) supervised fine-tuning (SFT) on a blend of existing tool-calling training datasets, (2) SFT on a blend including the \dataset{} training dataset, and (3) SFT on existing tool-calling datasets followed by RPO \citep{nvidia2024nemotron4340btechnicalreport} on \dataset{}, described below. In each case, the LM is trained on a combination of generic datasets along with the tool-calling specific datasets to maintain overall capabilities like instruction following, chat ability, question-answering, knowledge-intensive, tasks, etc.

\subsubsection{Supervised Fine-Tuning} \label{sec:sft-training}
For tool-calling SFT, we use publicly available datasets \citep{liu_apigen_2024,glaiveai_glaive-function-calling-v2_2024} and sample them to maintain a balance between examples containing single tool-call generation, multiple tool-calls generation and generating answers from tool responses in a multi-turn conversation. As seen from the results in Table \ref{tab:training-results}, models trained on these datasets have significant room for improvement on the \dataset{} benchmark.

We apply the pipeline described in \ref{sec:data-pipeline} to the APIGen dataset to create fine-tuning data by using the correct answer choice as the target. This data is combined with the tool-calling SFT data described above to create the final \dataset{}-SFT data blend. A 2:1 ratio of examples involving tool-calling and examples involving "cannot answer" or "request for information" gave the best overall results in our experiments. A constant learning rate of 5e-6 for the 4B model and 4e-6 for the 8B model was used with no warm-up.

\subsubsection{Preference Optimization} \label{sec:dpo-training}

We also leverage the multiple-choice format of \dataset{} to create a preference dataset for RPO, where we provide the correct answer as the chosen response, and one incorrect answer as the rejected response. For each type of correct answer, we uniformly sample the incorrect answer out of direct answer, tool call, follow-up question, and ``unable to answer'' categories. To prevent regression on tool-calling ability, we also add a subset where each chosen response is a tool-call, and the rejected response is created by either (1) removing required parameters from the correct tool-call, (2) modifying the tool-call arguments to have incorrect values, or (3) dropping a subset of tool-calls when the correct response contains more than one. The final dataset is created by combining these two subsets in a 1:1 ratio. We find that a low KL-penalty value (0.05) gives the best results on tool-calling benchmarks with this dataset. We use a constant learning rate of 9e-7 and 7e-7 for the 4B and 8B models, respectively, with a warm-up of 10 steps.

\subsection{Results from SFT \& RPO}

\Cref{tab:training-results} shows the results of SFT and RPO on \dataset{} compared to a baseline tool-calling SFT blend.
Firstly, we find that a targeted blend of existing tool-calling datasets that maintains diversity in tools and balances tool-calling examples with examples that do not involve tool-calling goes a long way, with our baseline SFT models already outperforming community models both in their size class and beyond on \dataset{}, as well as performing competitively on BFCL. 
Secondly, we find that while adding \dataset{} to the SFT blend improves results on \dataset{}, it results in a 6.2\% drop on BFCL Live AST for the 4B model, causing the model to become a little too conservative. This is ameliorated by doing RPO training instead of SFT, which yields a smaller drop on BFCL Live AST for the 4B model and yields an increase on all datasets for the 8B model. 
Collectively, these results highlight the importance of curating a targeted dataset blend and training regime to ensure an optimal trade-off between calling tools when possible and being conservative when not.

\section{LLM-as-Judge Evaluation} \label{sec:llm-as-judge}

One limitation of evaluating our multiple-choice benchmark with log-probabilities is that we cannot evaluate closed-source tool-calling models. Thus, we implement a second, alternative LLM-as-judge metric.
By comparing the two metrics, we can also alleviate concerns that the precise answer phrasing chosen by \genmodel{} when generating our benchmark might affect model performance.%

This method prompts the target LM exactly the same as for multiple-choice but has the LM generate a free-form answer instead of getting log-probabilities.
We then use an LLM-as-judge, specifically GPT-4-Turbo-04-09, to classify the target LM's generated output into the four multiple-choice categories: direct answer, tool call, follow-up question and unable to answer.  %
Since we only ask the LLM-as-judge to classify the output among these categories, we are not dependent on the judge's own tool-calling (or when-to-call) capability.
Then, we can calculate the category accuracy (now direct, not normed by length) and F1 score for the multiple-choice version. Tool hallucination rate over the questions where no tool is provided can be calculated in the same way as before.
We use 100 representative examples from each category of \dataset{} to keep costs accessible, as shown in \Cref{tab:statistics}.

\begin{table*}[t]
    \centering
    \begin{tabular}{l ccc cc} \toprule
        \textbf{Model} & \multicolumn{3}{c}{\textbf{\dataset{}}} & \textbf{BFCL AST} & \textbf{BFCL Irr.} \\
         & F1 $\uparrow$ & Acc $\uparrow$  & Tool Hall\% $\downarrow$ & Acc $\uparrow$  & Acc $\uparrow$   \\
        \midrule
        GPT-4o & 61.3 & 61.3\% & 26\% & \textbf{79.8\%} & \textbf{83.8\%}\\
        GPT-4o-Mini & 52.9 & 54.2\% & 41\% & 76.5\%  & 80.7\%  \\
        GPT-4-Turbo-04-09 & \textbf{64.6} & \textbf{64.3\%} & \textbf{22\%} & 63.8\%  & 35.6\%  \\
    \bottomrule
    \end{tabular}
    \caption{Results on \dataset{}, BFCL v2 Live AST and BFCL Irrelevance for three closed-source tool-calling models using LLM-as-judge evaluation. For BFCL, we report the scores using prompting, not native function-calling, for best comparison. The \textbf{best} scores are bolded; see \Cref{tab:training-results} for comparison with community models.
    }
    \label{tab:closed-source-results}
\end{table*}

\subsection{Comparison with Multiple-Choice Results}

\Cref{tab:llm-as-judge-comparison} shows the comparison in F1 scores for our 4B and 8B models under the multiple-choice with log-probabilities and LLM-as-judge evaluation methods. 
The two methods yield similar performance for the baseline models, which are trained only on open-source datasets, but sometimes underestimate performance for our \dataset{}-trained SFT and RPO models, which see their own non-tool-call answer phrasings during training and thus appear to be more affected by the specific phrasing of the multiple-choice answers.
Thus, we recommend using the LLM-as-judge method where possible and affordable for models that are trained on how to answer when \textit{not} calling tools, but results appear to be similar for models primarily trained on datasets that only evaluate correct tool-calling.

\begin{table}[t]
    \centering
    \begin{tabular}{l c c} \toprule
        \textbf{Model} & \textbf{MC} & \textbf{LLM-as-Judge} \\
         & F1 & F1  \\
        \midrule
        MNM 4B baseline & 29.7 & 27.9 \\
        MNM 4B SFT & 48.1 & 48.6 \\
        MNM 4B RPO & 51.0 & 64.3 \\
        MNM 8B baseline & 31.9 & 34.8 \\
        MNM 8B SFT & 49.4 & 57.1 \\
        MNM 8B RPO & 52.4 & 66.1 \\
    \bottomrule
    \end{tabular}
    \caption{Results on \dataset{} for multiple-choice vs. LLM-as-judge evaluation. F1 scores are comparable for models trained only on tool-calling datasets, but multiple-choice sometimes underestimates performance for models that see specific answer phrasings as part of the \dataset{} training set.
    }
    \label{tab:llm-as-judge-comparison}
\end{table}

\subsection{Results for closed-source models}

We evaluate three GPT-4 models using the LLM-as-judge method since closed-source tool-calling models are often reported to be better than open-source variants (\citealp{zhang_toolbehonest_2024} i.a.). Results are shown in \Cref{tab:closed-source-results}. %
Indeed, we find that the GPT-4 models outperform the community models in \Cref{tab:training-results}, including the 4B and 8B models fine-tuned on \dataset{}. Nonetheless, there is still room for improvement. In particular, their tool hallucination rates are not better than the community models.

\section{Discussion}

\paragraph{Improving when-to-call accuracy is not trivial}
We find that improving scores on when-\emph{not}-to-call benchmarks like BFCL Irrelevance and \dataset{} is not as simple as training on negative examples, as this can make the model over-conservative about calling tools and cause a dip in performance on BFCL AST.
For example, simply adding negative examples of tool-calling to the instruction-tuning blend of our 4B model by randomly pairing tool specifications with unrelated instruction-following questions in the SFT blend increases performance on \dataset{} %
but decreases performance on BFCL AST, as the model becomes too conservative and does not call tools often enough.
A similar issue arises if we simply add the entire \dataset{} training set to our instruction-tuning blend: performance improves on \dataset{}, but the model becomes too conservative and drops slightly in performance on BFCL (\Cref{sec:sft-training}).

This issue is compounded if SFT is followed by a helpfulness training step.
Instead, we propose using an appropriately balanced sample of the \dataset{} training set with RPO training (\Cref{sec:dpo-training}), which successfully balances these two competing pressures and enhances the training signal by providing negative as well as positive examples.

\paragraph{How does \dataset{} compare to BFCL Irrelevance?}

\begin{figure}[t]
\includegraphics[width=0.49\textwidth]{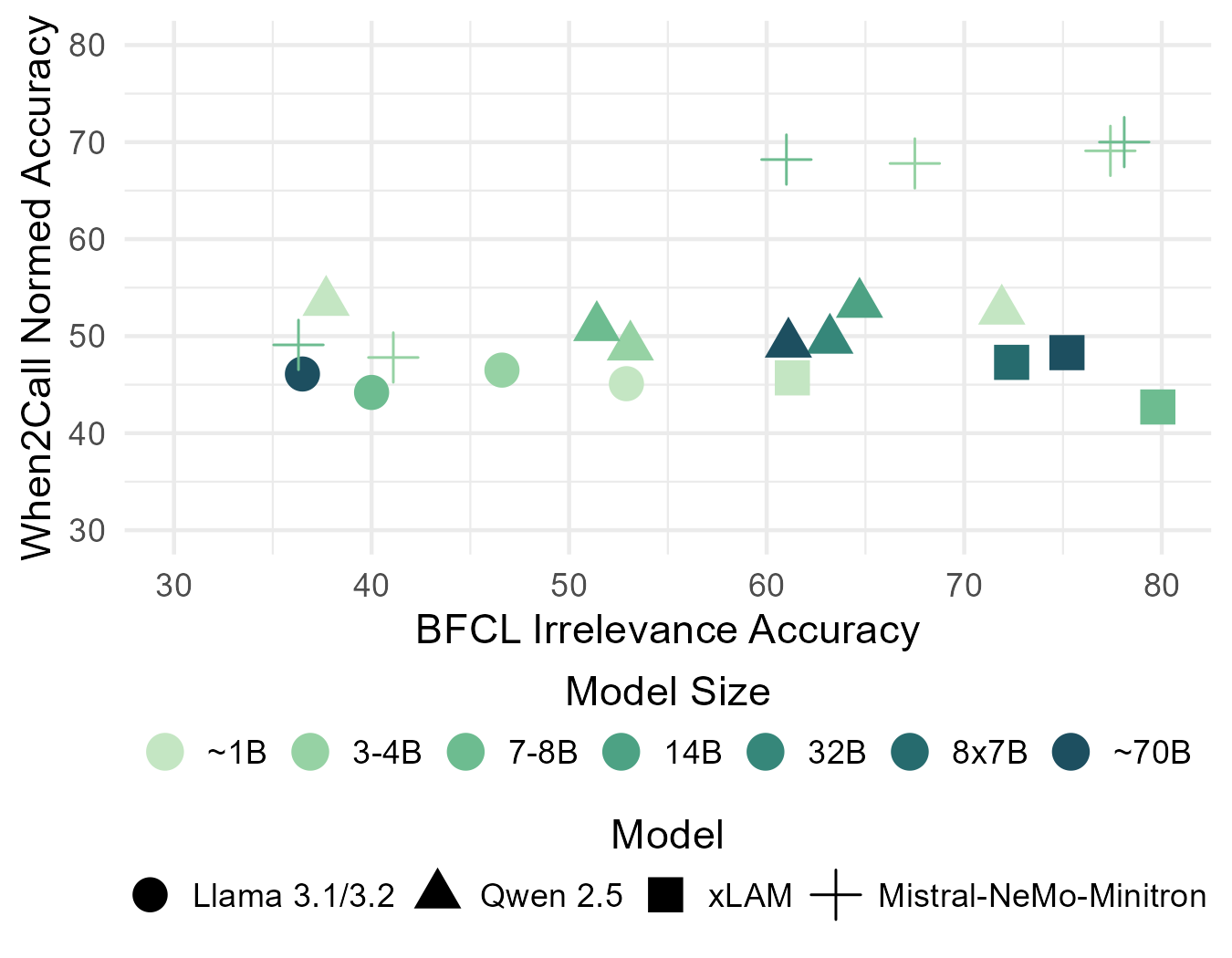}
    \caption{\dataset{} measures more complex capabilities than BFCL Irrelevance: a high score on BFCL Irrelevance need not yield a high score on \dataset{}, which indicates that \dataset{} offers a more fine-grained and more challenging task. 
    }\label{fig:irrelevance-when2call-scatter}
\end{figure}

As shown in \Cref{fig:irrelevance-when2call-scatter}, \dataset{} does not measure the same question as BFCL Irrelevance. Good performance on BFCL Irrelevance, i.e. \emph{not} calling a highly mismatching tool, is not enough to predict good performance on \dataset{}, which firstly requires the model to choose the correct other behavior (asking a follow-up question or admitting it can't answer), and secondly has subtler mismatches between the questions and the provided tools (\Cref{sec:tool-mismatch-difficulty}). This is a finer-grained and more difficult task, which is often largely outside the models' training data, even when they are trained on data mimicking BFCL Irrelevance.

\paragraph{Do models ask follow-up questions when required to call the tool?}

A novel contribution of this dataset is explicitly testing whether models can ask follow-up questions when the tool matches the question but the required information is missing. 
As illustrated in the confusion matrices in \Cref{sec:confusion-matrices}, we find that the Qwen, xLAM, and fine-tuned Mistral-NeMo-Minitron models are able to correctly ask follow-up questions over half the time, though they still often hallucinate a tool call with the missing parameters instead. 

\paragraph{Is performance on \dataset{} low because models always call tools?}

One might expect that performance on \dataset{} is low because models trained on data like APIGen \citep{liu_apigen_2024} are insufficiently conservative and call tools too often, having rarely seen examples of ``unable to answer'' or follow-up questions in relation to tools. In fact, the confusion matrices in \Cref{sec:confusion-matrices} show that only the Llama models do this. The Qwen, xLAM, and Mistral-NeMo-Minitron models
each have their own error pattern, often correctly not calling a tool in many cases but preferring an incorrect answer among the remaining three text options. Qwen and xLAM models are highly unwilling to select the (d) ``unable to answer'' option, selecting either a direct answer or a request for information instead. This results in a low macro F1 score since the F1 score for (d) is so low, even though their (micro-averaged) accuracy may be relatively high.  

\paragraph{\dataset{} measures tool, answer, and parameter hallucination}

\dataset{} allows developers to measure each type of possible hallucination (tool, direct answer, missing information) and adjust the system prompt or training regime depending on the type of errors the model is making, in complement with BFCL's evaluation of the accuracy of the tool calls. For example, the RPO training blend or answer pairs (\Cref{sec:dpo-training}) can be adjusted to demonstrate the correct trade-offs.
Answer hallucination and parameter hallucination rates can be read directly off the confusion matrix, while we provide a script to calculate tool hallucination -- see \Cref{sec:hallucination-calculation}.

\section{Related Work} \label{sec:related-work}

With tool-calling growing in popularity (see \citet{qu_tool_2024} for a survey), a number of tool-calling benchmarks have been developed.
While some benchmarks break the tool-calling process into subtasks \citep{li-etal-2023-api,basu-etal-2024-api,ye_tooleyes_2024,huang_metatool_2024},
most recent benchmarks treat tool-calling as a single step, a trend which our benchmark follows. In this case, the model must return either a tool call or an appropriate text response.
However, most benchmarks only test the case where the correct API is provided and focus either on validating the tool call \citep{patil_gorilla_2023,xu_tool_2023,nexusflow_nexus_2023} or evaluating the final answer or system state, which assumes a correct tool call \citep{yang_gpt4tools_2023,tang_toolalpaca_2023,qin_toolllm_2023,zhuang_toolqa_2023,guo_stabletoolbench_2024,yao_tau-bench_2024}.

Among benchmarks, only BFCL \citep{yan_berkeley-v1_2024} and ToolSandbox \citep{lu_toolsandbox_2024} attempt to address cases where the correct API is not provided, or information is missing with their Irrelevance and Insufficient Information categories respectively, which provide no correct tool. ToolSandbox and BFCL's Multi-Turn category also contain some examples of follow-up questions. Both of these categories, however, only evaluate whether a tool call is made or not, and not what the model does instead. %
ToolBeHonest \citep{zhang_toolbehonest_2024} also considers these cases but only evaluates the subtask of whether the task is solvable. 
For training data, the Glaive v2 training dataset \citep{glaiveai_glaive-function-calling-v2_2024} is the only publicly available training dataset to include cases where the correct API is not provided, or information is missing, but it does not separate such items from the rest of the training set, or provide an evaluation metric other than exact match / cross-entropy loss. Other popular training datasets like Hermes \citep{interstellarninja_hermes_2024} and APIGen \citep{liu_apigen_2024} do not cover these cases.

Tool hallucinations have only been tackled very recently. \citet{abdelaziz_granite-function_2024} report the tool hallucination rate of their model, Granite. The ToolBeHonest benchmark \citep{zhang_toolbehonest_2024} studies tool hallucination with an indirect measure by asking models to classify the solvability of tasks (``solvable'' indicates that the model is hallucinating that the tool can be used). While the hallucination rate can be calculated manually from the detailed output of BFCL, no previous benchmark provides explicit support for calculating tool hallucination rate, and none study how often the model hallucinates a text answer instead.

\section{Conclusion}
We presented \dataset{}, a new synthetically generated dataset to evaluate when tool-calling LMs should (not) call tools and how they should behave if they can't, using a multiple-choice format over four different types of behavior. We present a traditional accuracy/F1 metric using log-probabilities as well as an LLM-as-judge alternative which allows the evaluation of generated outputs, particularly of closed-source models.
Unlike the main previous benchmark evaluating when not to call tools, BFCL Irrelevance, which is already beginning to saturate\footnote{\scriptsize \url{https://gorilla.cs.berkeley.edu/leaderboard.html}}, 
we find that even large tool-calling models still struggle on our more difficult task, which requires not just \emph{not} calling a tool, but choosing the correct non-tool-call response, and has a higher similarity between the question and the nonetheless incorrect tool calls. 
Further, we show that achieving the right degree of conservativeness for tool-calling models is not trivial, as simple instruction-tuning on the \dataset{} training dataset in conjunction with other instruction-following and helpfulness data can lead to over-conservative behavior. We propose an RPO training method that leverages the multiple-choice nature of the dataset and strikes a balance between the pressures of when vs. when not to call tools.
Training on \dataset{} may also lead the model to overly prefer specific answer phrasings, however, impacting its score using log-probability multiple choice method. We recommend using the LLM-as-judge method, where possible, after training on \dataset{} to mitigate this.
Finally, we offer scripts to calculate the confusion matrix and hallucination rate as part of model evaluation, allowing model developers to understand the individual failure patterns of their model beyond a single accuracy score and develop a targeted training regime in response.

\section*{Limitations} \label{sec:limitations}

\paragraph{Quality limitations}
Since we are using a synthetic data generation pipeline, some dataset quality issues remain despite multiple iterations of prompt tuning.
In \Cref{sec:quality}, we discuss how we evaluated dataset quality and what issues remain. The overall quality percentage (manually estimated on a subset of the data) is 92\% for questions and 94\% for question answers.

\paragraph{Assumption that the direct answer is incorrect}
In order to have a clear, correct answer, \dataset{} is constructed to assume that the direct answer is always a hallucination and, thus, always wrong. 
However, this simplifying assumption only reflects one kind of real-world tool use. In other cases, especially for small language models deployed on devices, the task may be solvable without a tool in principle, but the LM may wish to avail itself of a tool anyway in order to improve performance. A classic example of this is mathematics and calculation questions, which form a small part of BFCL Live and which we filter out when generating our dataset.
These tasks present a trade-off between compute and accuracy: calling a tool will increase accuracy at the expense of compute and response time.
Any future benchmark that covers such tasks will need to be flexible enough that users with different preferences for this trade-off can interpret or adjust the benchmark accordingly.

\paragraph{Evaluation of closed-source models}
We were only able to evaluate closed-source models from the GPT family for this paper. Other closed-source models also show good performance on BFCL. We hope to evaluate additional models for a future edition of the benchmark.

\paragraph{Languages used}
Like previous tool-calling work, we focus on English-language questions and answers only. Expanding tool-calling to other languages is certainly an important research direction; we hope to see multilingual analogs of BFCL in the near future.

\bibliography{anthology,custom}

\appendix

\section{Quality checklist for synthetic data generation} \label{sec:quality}

Synthetically generated data can be noisy, so human verification is important. We manually inspected ca. 10\% of the generated dataset (50 samples per type) for quality at each iteration of prompt tuning, and adjusted the prompt accordingly.
\Cref{tab:quality-checklist} shows the checklist we used and the results on the final iteration of the dataset. The checklist was generated by manually inspecting early versions of the dataset and listing all the errors that were observed. 
The overall answer quality percentage (out of 150 synthetic answers) is 94\%. The overall question quality percentage (out of 100 synthetic questions) is 82\%. 

One particular issue is that some questions inherited from BFCL and classified as requiring tool use do need a tool to be answered properly, but also admit vague partial answers. (One example asks for a specific forecast of tree growth in Yellowstone National Park, which can be partially answered by replying that the trees will grow a moderate amount.) Sometimes, our pipeline generates such a vague answer as its direct answer option (a), meaning that (a) is not entirely incorrect as we assume it to be. While it is likely still a better answer for the model to say it can't answer the exact question than to be vague, this is now a more subjective judgment. This error arises in part from the sometimes unusually specific tools (and corresponding questions) that BFCL employs, which only require tool use in that exact formulation, and might not all represent realistic tools in the wild.

Two other issues relate to the ``unable to answer'' option.
Sometimes, the generating model may generate an answer intended to be the ``unable to answer'' option (d), but which also contains a follow-up question.
Further, when we ask the model to generate a question that cannot be answered with the given tool (a difficult task), it very occasionally generates a question that can, in fact, be answered with the tool, though usually still not with the tool call we provide as answer (b). While ``unable to answer'' (d) is still the best choice of the four options, this may confuse the model as its most preferred answer would be to generate a valid tool call.

In future work, we plan to implement an additional LLM-as-judge filtering step which checks for these issues.

\begin{table*}[t]
    \centering
    \begin{tabular}{lrr}
        \toprule
        \textbf{Issue Type} & \textbf{Count} & \textbf{Percentage} \\
        \midrule
        \multicolumn{3}{l}{\textbf{Answer type: follow-up question}} \\
        Asks about output format & 0 & 0.0\% \\
        Asks to confirm already provided values & 3 & 6.0\% \\
        Asks for already provided information & 0 & 0.0\% \\
        Asks for reason or context & 0 & 0.0\% \\
        Asks about additional inputs to pass to the tool & 0 & 0.0\% \\
        Asks for something irrelevant & 0 & 0.0\% \\
        Hallucinates other information & 0 & 0.0\% \\
        \midrule
        \textbf{Total} & 3 & 6.0\% \\
        \midrule\midrule
        \multicolumn{3}{l}{\textbf{Answer type: direct answer}} \\
        Contains request for information & 0 & 0.0\% \\
        \midrule
        \textbf{Total} & 0 & 0.0\% \\
        \midrule\midrule
        \multicolumn{3}{l}{\textbf{Answer type: unable to answer}} \\
        Mentions information not included in question  & 0 & 0.0\% \\
        \midrule
        \textbf{Total} & 0 & 0.0\% \\
        \midrule\midrule
        \multicolumn{3}{l}{\textbf{Question type: correct answer is ``unable to answer''}} \\
        Includes explanation referencing tool capabilities & 0 & 0.0\% \\
        Answerable with provided tool & 0 & 0.0\% \\
        Generic / vague terms (no specific values) & 2 & 4.0\% \\
        References things that are not mentioned (using ``the'') & 4 & 8.0\% \\
        Totally unrelated to tool & 0 & 0.0\% \\
        Question doesn't need a tool call to answer & 0 & 0.0\% \\
        Question partially answerable without a tool & 3 & 6.0\% \\
        \midrule
        \textbf{Total} & 9 & 18.0\% \\
        \midrule\midrule
        \multicolumn{3}{l}{\textbf{Question type: correct answer is follow-up question}} \\
        Does not have any missing information & 0 & 0.0\% \\
        Mentions vague parameter values & 0 & 0.0\% \\
        Says that a value is not provided & 0 & 0.0\% \\
        References existence of tool & 0 & 0.0\% \\
        \midrule
        \textbf{Total} & 0 & 0.0\% \\
\bottomrule
\end{tabular}
\caption{Quality checklist for synthetically generated questions and answers. Overall answer quality percentage (out of 150 synthetic answers): 94\%. Overall question quality percentage (out of 100 synthetic questions): 82\%.}\label{tab:quality-checklist}
\end{table*}

\section{System prompts in \dataset{}} \label{sec:system-prompts}

\Cref{tab:default_prompt} shows the default system prompt that we use in \dataset{} for models that do not come with a pre-specified system prompt and/or tool-calling format. \Cref{tab:qwen_prompt} shows how we incorporate existing system prompts for models, using Qwen as an example. The prompt is taken verbatim from the documentation for Qwen 2.5.\footnote{\url{https://qwen.readthedocs.io/en/latest/framework/function_call.html}}
The placeholders \texttt{tool} / \texttt{tools} and \texttt{question} indicate where the provided tools and question are included in the prompt.

\begin{table*}[t]
    \centering
    \footnotesize
    \begin{adjustbox}{width=\textwidth}
    \begin{tabular}{p{\linewidth}}
        \toprule
\begin{verbatim}You are a helpful AI assistant. 
You have access to the tools described in <tool></tool> which you can use to answer the user's questions.
Only use a tool if it directly answers the user's question.

To use a tool, return JSON in the following format:
{"name": "tool_name", "arguments": {"argument1": "value1", "argument2": "value2", ...}}

<tool>{tool}</tool>
<tool>{tool}</tool>
...

{question}
\end{verbatim}
\\\bottomrule
\end{tabular}
\end{adjustbox}
\caption{Prompt used in \dataset{} to evaluate models that don't have their own system prompt. We use a minimalist prompt since we do not want to give these models an advantage over other models whose pre-specified system prompt does not include any information about when \textit{not} to call tools.}\label{tab:default_prompt}
\end{table*}

\begin{table*}[t]
    \centering
    \footnotesize
    \begin{adjustbox}{width=\textwidth}
    \begin{tabular}{p{\linewidth}}
        \toprule
\begin{verbatim}<|im_start|>system
You are Qwen, created by Alibaba Cloud. You are a helpful assistant.

# Tools

You may call one or more functions to assist with the user query.

You are provided with function signatures within <tools></tools> XML tags:
<tools>
{tools}
</tools>

For each function call, return a json object with function name and arguments within 
<tool_call></tool_call> XML tags:
<tool_call>
{{"name": <function-name>, "arguments": <args-json-object>}}
</tool_call><|im_end|>
<|im_start|>user
{question}<|im_end|>
<|im_start|>assistant
\end{verbatim}\\
\\\bottomrule
\end{tabular}
\end{adjustbox}
\caption{Prompt used in \dataset{} to evaluate Qwen 2.5 models, which specifies a custom system prompt to use for tool-calling. We use each model's preferred prompt in order to match their training/fine-tuning conditions.}\label{tab:qwen_prompt}
\end{table*}

\section{Prompts for synthetic data generation} \label{sec:synth-prompts}

\begin{table*}[t]
    \centering
    \footnotesize
    \begin{adjustbox}{width=\textwidth}
    \begin{tabular}{p{\linewidth}}
        \toprule
\begin{verbatim}Questions can be answered in the following ways:
- Using public data, available from books or internet datasets
- Using a calculator and/or mathematical or physical formulas
- Using a specialized tool, such as statistical software, music software or machine learning libraries
- Using real-time information, such as weather, stock prices or up-to-date ratings
- Using databases (private or public), such as access to player statistics, customer records or lawsuits

Here are some examples:

Question: What is the weather in London tomorrow?
Category: Real-time information

Question: What is the specific heat capacity of water?
Category: Public data

Question: What is the length of the hypotenuse of a right triangle with side lengths 4 and 3?
Category: Calculator

Question: Find all lawsuits in New York State between 2010-2012.
Category: Database

Question: Generate a melody in C major.
Category: Specialized tool

Question: How long does it take to drive from Boston to New York?
Category: Real-time information

Question: Find popular Indian restaurants in Las Vegas.
Category: Real-time information

Question: What is the atomic number of oxygen?
Category: Public data

Question: Get the current level of my character in The Legend of Zelda: Breath of the Wild.
Category: Database

Question: What are the opening hours of Walmart in Santa Clara, CA?
Category: Real-time information

Question: What is the magnetic field strength 1 meter away from a wire with a 2 Ampere current?
Category: Calculator

Question: Perform a Chi-Squared test for independence on a 2x2 contingency table [[1, 2], [3, 4]]
Category: Specialized tool

Now, classify this question into one of "Public data", "Calculator", "Specialized tool",
"Real-time information", or "Database", using the format "Category: <category>":

Question: {question}
\end{verbatim}
\\\bottomrule
\end{tabular}
\end{adjustbox}
\caption{Prompt template used for tool use classification.}
\label{tab:classification_prompt}
\end{table*}

\begin{table*}[t]
    \centering
    \footnotesize
    \begin{adjustbox}{width=\textwidth}
    \begin{tabular}{p{\linewidth}}
        \toprule
Question generation where the correct answer is (c), follow-up question \\
\midrule
\begin{verbatim}You are given a question and a tool specification in json format.
The tool can be used to answer the provided question.
The tool has certain parameters that are required to use it.
The question provides a value for each of these parameters.

- Your task is to re-write the question such that it does not proide any value 
for the `{required_param_to_remove}` parameter.
- The re-written question must be consistent with original question in meaning
- All other provided values must remain the same, except the one to be excluded.
- There should not be any mention of the excluded parameter in the question.
- The question should not use phrases like "a specific location" or "a specific date".
It should omit this information entirely.

Respond only with the re-written question and nothing else.
Here is the original question and parameter to remove -
Original question: {original_question}
Parameter to remove from question: {required_param_to_remove}
\end{verbatim}\\\\\midrule\midrule

Question generation where the correct answer is (d), unable to answer \\
\midrule
\begin{verbatim}You are an expert at writing technical content.

[tool] {tool} [/tool]

- The tools mentioned above inside [tool] [/tool] can be used to answer certain questions.
- Give one example of a question that none of these tools can be used to answer.
- The question should ask about a specific case and provide all relevant information.
- Give specific numbers and values where applicable.
- The question should be one complete sentence and include a question mark.
- The question should be no more than 10-30 words.
- Specify all the necessary information, but otherwise keep the question short.
- Give one example question and nothing else.
- Do not explain why the question cannot be answered. Do not wrap the question in quotes.
\end{verbatim}

\\\bottomrule
\end{tabular}
\end{adjustbox}
\caption{Prompt templates used for generating questions in \dataset{} examples, as discussed in \Cref{sec:data-pipeline}. \label{tab:example_question_prompts}}
\end{table*}

\begin{table*}[t]
    \centering
    \footnotesize
    \begin{adjustbox}{width=\textwidth}
    \begin{tabular}{p{\linewidth}}
        \toprule
Response generation for category (c), where the response is a follow-up question \\
\midrule
\begin{verbatim}You are an expert at writing dialogues involving technical content.
Your task is to write a continuation to a conversation between a User and an Assistant.
The Assistant has access to the following tool which can be used to answer User queries:
[tool] {tool} [/tool]

The User will ask a question to the Assistant.
You must write the Assistant's response to this question by following the instructions given below:
- The User's question does not provide `{removed_param}` which is a required parameter to use the tool.
- Assistant requires some additional information to answer the question or to use the provided tool.
- The Assistant should ask for `{removed_param}` from the User.
- The Assistant should not ask to clarify or confirm any information that the User already provided.
- The Assistant's question should not be about the answer format or any other formatting.
- The Assistant's question should be no more than 50 words (shorter is fine). 
- Stop after generating the Assistant's query for more information. Do not generate a tool call.
- Do not include a word count or any information regarding word count in your answer.
- Do not provide a note or any other content in your response. Respond only with the Assistant's reply.

Here is the conversation so far -
User: {rewritten_question}
Assistant:
\end{verbatim}\\\\\midrule\midrule

Response generation for category (d), where the response is "unable to answer" \\
\midrule
\begin{verbatim}You are an expert at writing dialogues involving technical content.
Your task is to write a continuation to a conversation between a User and an Assistant.

The User will ask a question to the Assistant.
You must write the Assistant's response to this question by following the instructions given below:
- Assume that the Assistant does not know the answer, even if you know the answer.
- The Assistant should explain that it can't answer the question as it cannot perform the requested task.
- The Assistant's answer should be no more than 40 words (shorter is fine).
- Stop after generating the Assistant's answer. Do not generate the User's response.
- Do not include a word count or any information regarding word count in your answer.
- Do not provide a note or any other content in your response. Respond only with the Assistant's reply.

Here is the conversation so far -
User: {question}
Assistant:
\end{verbatim}\\\\\midrule\midrule

Response generation for category (a), where the response is a direct answer without any tool use \\
\midrule
\begin{verbatim}You are an expert at writing dialogues involving technical content.
Your task is to write a continuation to a conversation between a User and an Assistant.

The User will ask a question to the Assistant.
You must write the Assistant's response to this question by following the instructions given below:
- Assume that the Assistant knows the correct answer to the question. Do not ask follow-up questions.
    If necessary, the Assistant should guess any missing information.
- Keep the answer simple. Do not provide disclaimers about accuracy.
- The Assistant's answer should be no more than 50 words, if possible (shorter is fine 
    if the answer is simple).
- Stop after generating the Assistant response.
- Do not include a word count or any information regarding word count in your answer.
- Do not provide a note or any other content in your response. Respond only with the Assistant's reply.

Here is the conversation so far -
User: {question}
Assistant:
\end{verbatim}
\\\bottomrule
\end{tabular}
\end{adjustbox}
\caption{Prompt templates used for generating responses used as choices in \dataset{} examples, as discussed in \Cref{sec:data-pipeline}. \label{tab:example_response_prompts}}
\end{table*}

The prompt template used for tool use classification is shown in \Cref{tab:classification_prompt}. 
The prompts for synthetic data generation are shown in \Cref{tab:example_question_prompts}.

\section{Results for all models}

Table \ref{tab:small-model-results} the results table with all models included, including some small and intermediate model sizes omitted in Table \ref{tab:training-results}.

\begin{table*}[t]
    \centering
    \begin{tabular}{l ccc cc} \toprule
        \textbf{Model} & \multicolumn{3}{c}{\textbf{\dataset{}}} & \textbf{BFCL AST} & \textbf{BFCL Irr.} \\
         & F1 $\uparrow$ & Acc-Norm $\uparrow$  & Tool Hall\% $\downarrow$ & Acc $\uparrow$  & Acc $\uparrow$   \\
        \midrule
        Llama 3.2 1B Instruct & 21.7 & 45.1\% & 43\% & 13.2\% & 52.9\% \\
        Llama 3.2 3B Instruct & 17.9 & 46.5\% & 52\% & 37.6\% & 46.6\% \\
        Llama 3.1 8B Instruct & 16.6 & 44.2\% & 67\% & 51.6\% & 40.0\% \\
        Llama 3.1 70B Instruct & 37.8 & 46.1\% & 57\% & \underline{68.3}\% & 36.5\% \\
        Qwen 2.5 0.5B Instruct & 32.0 & 53.5\% & 20\% & 22.9\% & 37.7\% \\
        Qwen 2.5 1.5B Instruct & 29.9 & 52.6\% & 23\% & 36.5\% & 71.9\% \\
        Qwen 2.5 3B Instruct & 29.8 & 48.9\% & 23\% & 54.8\% & 53.1\% \\
        Qwen 2.5 7B Instruct & 32.0 & 50.9\% & 21\% & 64.1\% & 51.4\% \\
        Qwen 2.5 14B Instruct & 36.2 & 53.3\% & 21\% & 61.6\% & 64.7\% \\
        Qwen 2.5 32B Instruct & 32.9 & 49.6\% & 17\% & 65.6\% & 63.2\% \\
        Qwen 2.5 72B Instruct & 32.8 & 49.2\% & 23\% & \textbf{69.3\%} & 61.1\% \\
        xLAM 1B FC-R & 25.6 & 45.7\% & 40\% & 55.3\% & 61.3\% \\
        xLAM 7B FC-R & 31.5 & 42.7\% & 24\% & 58.3\% & \textbf{79.8\%} \\
        xLAM 8x7B R & 32.9 & 47.3\% & 13\% & 67.5\% & 72.4\% \\
        xLAM 8x22B R & 34.3 & 48.3\% & 9.0\% & 74.7\% & 75.2\% \\
        \midrule
        MNM 4B SFT (baseline) & 29.7 & 47.8\% & 16\% & 57.9\% & 41.1\% \\
        MNM 4B \dataset{}-SFT & 48.1 & 67.8\% & 4.3\% & 51.7\% & 67.5\% \\
        MNM 4B \dataset{}-RPO & \underline{51.0} & \underline{69.1\%} & \underline{1.9\%} & 54.0\% & 77.4\% \\
        MNM 8B SFT (baseline) & 31.9 & 49.1\% & 19\% & 62.2\% & 36.3\% \\
        MNM 8B \dataset{}-SFT & 49.4 & 68.2\% & 7.0\% & 57.5\% & 61.0\% \\
        MNM 8B \dataset{}-RPO & \textbf{52.4} & \textbf{70.0\%} & \textbf{1.2\%} & 62.5\% & \underline{78.1\%} \\
    \bottomrule
    \end{tabular}
    \caption{Results on \dataset{}, BFCL Live AST and BFCL Irrelevance for community tool-calling models, and for our Mistral-NeMo-Minitron models with and without training on \dataset{} using SFT and RPO. 
    For \dataset{}, we show Macro F1, length-normed accuracy and the tool hallucination rate when no tools are provided (lower is better $\downarrow$; see \Cref{sec:hallucination-calculation} for calculation).
    Models not trained on \dataset{} struggle to make the right choices; RPO training yields the greatest benefits. The \textbf{best} and \underline{second-best} scores are bolded and underlined.
    Additional model sizes in \Cref{tab:small-model-results}.
    }
    \label{tab:small-model-results}
\end{table*}

\section{Confusion matrices and hallucination rates}

\subsection{Calculating hallucination rates} \label{sec:hallucination-calculation}

We design \dataset{} so that we can directly calculate the proportions of each type of hallucination.
Answer hallucination occurs whenever the model chooses the direct answer (a), and can be read directly off the confusion matrix. We provide a script to generate the confusion matrices; see \Cref{sec:confusion-matrices} for examples. 
Tool hallucination can be evaluated using the questions where no tools are provided at all. If a model chooses the tool call answer in this scenario, they are necessarily hallucinating the tool. 
We provide a script to calculate this rate directly.
Finally, parameter hallucination can also be read off the confusion matrix by counting the questions where information was missing (i.e., a follow-up question (c) was correct) where the model chose the tool call answer (b).

\subsection{Confusion matrices on \dataset{}}\label{sec:confusion-matrices}

We provide the confusion matrices for selected models to illustrate that models may have different error patterns, even while achieving similar accuracies or similar F1 scores (a combination of accuracy and F1 partially reflects these differences). 
Tables \ref{tab:confusion-matrix-llama-70b}, \ref{tab:confusion-matrix-qwen-7b}, \ref{tab:confusion-matrix-qwen-72b} and \ref{tab:confusion-matrix-xlam-7b} show the confusion matrices for Llama 3.1 70B Instruct, Qwen 2.5 7B Instruct, Qwen 2.5 72B Instruct and xLAM 7B FC-R respectively. Tables \ref{tab:confusion-matrix-8b-baseline}, \ref{tab:confusion-matrix-8b-sft} and \ref{tab:confusion-matrix-8b-rpo} show the confusion matrices for the three versions of our Mistral-NeMo-Minitron 8B model.

\begin{table*}[t]
    \centering
    \begin{tabular}{lrrrr} \toprule
        & \multicolumn{4}{c}{\textbf{Predicted}} \\
        \textbf{True} & Direct answer & Tool call & Follow-up question & Unable to answer \\
        \midrule
        Direct answer & \textbf{0} & 0 & 0 & 0 \\
        Tool call & 0 & \textbf{1287} & 5 & 0 \\
        Follow-up question & 4 & 1009 & \textbf{44} & 5 \\
        Unable to answer & 100 & 1030 & 115 & \textbf{50} \\
    \bottomrule
    \end{tabular}
    \caption{Confusion matrix on \dataset{} for Llama 3.1 70B Instruct.}
    \label{tab:confusion-matrix-llama-70b}
\end{table*}

\begin{table*}[t]
    \centering
    \begin{tabular}{lrrrr} \toprule
        & \multicolumn{4}{c}{\textbf{Predicted}} \\
        \textbf{True} & Direct answer & Tool call & Follow-up question & Unable to answer \\
        \midrule
        Direct answer & \textbf{0} & 0 & 0 & 0 \\
        Tool call & 14 & \textbf{1078} & 201 & 2 \\
        Follow-up question & 16 & 410 & \textbf{634} & 2 \\
        Unable to answer & 119 & 453 & 645 & \textbf{78} \\
    \bottomrule
    \end{tabular}
    \caption{Confusion matrix on \dataset{} for Qwen 2.5 7B Instruct}
    \label{tab:confusion-matrix-qwen-7b}
\end{table*}

\begin{table*}[t]
    \centering
    \begin{tabular}{lrrrr} \toprule
        & \multicolumn{4}{c}{\textbf{Predicted}} \\
        \textbf{True} & Direct answer & Tool call & Follow-up question & Unable to answer \\
        \midrule
        Direct answer & \textbf{0} & 0 & 0 & 0 \\
        Tool call & 11 & \textbf{1116} & 167 & 1 \\
        Follow-up question & 12 & 365 & \textbf{684} & 1 \\
        Unable to answer & 162 & 478 & 599 & \textbf{56} \\
    \bottomrule
    \end{tabular}
    \caption{Confusion matrix on \dataset{} for Qwen 2.5 72B Instruct}
    \label{tab:confusion-matrix-qwen-72b}
\end{table*}

\begin{table*}[t]
    \centering
    \begin{tabular}{lrrrr} \toprule
        & \multicolumn{4}{c}{\textbf{Predicted}} \\
        \textbf{True} & Direct answer & Tool call & Follow-up question & Unable to answer \\
        \midrule
        Direct answer & \textbf{0} & 0 & 0 & 0 \\
        Tool call & 133 & \textbf{756} & 393 & 13 \\
        Follow-up question & 63 & 332 & \textbf{640} & 27 \\
        Unable to answer & 249 & 314 & 568 & \textbf{164} \\
    \bottomrule
    \end{tabular}
    \caption{Confusion matrix on \dataset{} for xLAM 7B FC-R}
    \label{tab:confusion-matrix-xlam-7b}
\end{table*}

\begin{table*}[t]
    \centering
    \begin{tabular}{lrrrr} \toprule
        & \multicolumn{4}{c}{\textbf{Predicted}} \\
        \textbf{True} & Direct answer & Tool call & Follow-up question & Unable to answer \\
        \midrule
        Direct answer & \textbf{0} & 0 & 0 & 0 \\
        Tool call & 7 & \textbf{1261} & 20 & 7 \\
        Follow-up question & 18 & 738 & \textbf{288} & 18 \\
        Unable to answer & 181 & 537 & 359 & \textbf{218} \\
    \bottomrule
    \end{tabular}
    \caption{Confusion matrix on \dataset{} for Mistral-NeMo-Minitron 8B SFT (baseline)}
    \label{tab:confusion-matrix-8b-baseline}
\end{table*}

\begin{table*}[t]
    \centering
    \begin{tabular}{lrrrr} \toprule
        & \multicolumn{4}{c}{\textbf{Predicted}} \\
        \textbf{True} & Direct answer & Tool call & Follow-up question & Unable to answer \\
        \midrule
        Direct answer & \textbf{0} & 0 & 0 & 0 \\
        Tool call & 11 & \textbf{1143} & 132 & 9 \\
        Follow-up question & 12 & 284 & \textbf{752} & 14 \\
        Unable to answer & 121 & 282 & 372 & \textbf{520} \\
    \bottomrule
    \end{tabular}
    \caption{Confusion matrix on \dataset{} for Mistral-NeMo-Minitron 8B SFT with \dataset{}}
    \label{tab:confusion-matrix-8b-sft}
\end{table*}

\begin{table*}[t]
    \centering
    \begin{tabular}{lrrrr} \toprule
        & \multicolumn{4}{c}{\textbf{Predicted}} \\
        \textbf{True} & Direct answer & Tool call & Follow-up question & Unable to answer \\
        \midrule
        Direct answer & \textbf{0} & 0 & 0 & 0 \\
        Tool call & 17 & \textbf{992} & 148 & 138 \\
        Follow-up question & 15 & 259 & \textbf{681} & 107 \\
        Unable to answer & 90 & 106 & 249 & \textbf{850} \\
    \bottomrule
    \end{tabular}
    \caption{Confusion matrix on \dataset{} for Mistral-NeMo-Minitron 8B RPO with \dataset{}}
    \label{tab:confusion-matrix-8b-rpo}
\end{table*}

\end{document}